\documentclass[11pt,twocolumn,twoside]{article}
\usepackage{fully3d}

\usepackage{amssymb,amsmath,siunitx}
\usepackage{tikz}
\usepackage{xfrac}
\usepackage{booktabs,tabularx}
\usepackage{makecell}
\newcolumntype{C}{>{\centering\arraybackslash}X}
\usepackage{enumitem}

\begin{document}

\title{Task-based Generation of Optimized Projection Sets using Differentiable Ranking} 

\author[1,2]{Linda-Sophie Schneider}
\author[1]{Mareike Thies}
\author[2]{Christopher Syben}
\author[2]{Richard Schielein}
\author[3]{Mathias Unberath}
\author[1,2]{Andreas Maier}

\affil[1]{\text{Pattern Recognition Lab},  Friedrich-Alexander-Universität Erlangen-Nürnberg, Germany }

\affil[2]{Fraunhofer Development Center X-Ray Technology EZRT, Germany}
\affil[3]{\text{Laboratory for Computational Sensing + Robotics}, Johns Hopkins University, Baltimore, MD, USA}

\maketitle
\thispagestyle{fancy}

\newcommand{\randlist}{} 

\newcommand{\getrandom}[1]{
	\pgfmathparse{random(0,#1)} 
	\pgfmathfloattofixed{\pgfmathresult} 
	\edef\temp{\pgfmathresult} 
	\ifx\temp\randlist 
	\expandafter\getrandom{#1} 
	\else
	\xdef\randlist{\randlist,\temp} 
	\temp
	\fi
}


\begin{customabstract}
	We present a method for selecting valuable projections in  computed tomography (CT) scans to enhance image reconstruction and diagnosis. The approach integrates two important factors, projection-based detectability and data completeness, into a single feed-forward neural network. The network evaluates the value of projections, processes them through a differentiable ranking function and makes the final selection using a straight-through estimator. Data completeness is ensured through the label provided during training. The approach eliminates the need for heuristically enforcing data completeness, which may excludes valuable projections. The method is evaluated on simulated data in a non-destructive testing scenario, where the aim is to maximize the reconstruction quality within a specified region of interest. We achieve comparable results to  previous methods, laying the foundation for using reconstruction-based loss functions to learn the selection of projections.
\end{customabstract}


\section{Introduction}
In the field of computed tomography (CT), a series of projections are obtained to produce a three-dimensional representation of the object of interest. However, not all projections are equally essential for image reconstruction and diagnostic purposes \cite{Thies}. A projection is considered more valuable when the amount of information gain in the chosen set of projections for reconstruction is higher. Selecting the most valuable projections can enhance the detection of anomalies or defects, improve imaging efficiency, and minimize noise and artefacts in the final reconstruction. In order to determine an optimized CT trajectory, it is necessary to balance the individual value of each projection with the overall value of the set of projections used for reconstruction.\\
 One approach to selecting valuable projections is through task-based image quality metrics. These metrics quantify the performance of a given projection with respect to a specific imaging task, such as detecting small structures or preserving low contrast details. By evaluating each projection using a task-based image quality metric, the most valuable projections can be selected for inclusion in the image reconstruction process. As a secondary criterion, data completeness should be satisfied. Simultaneously optimizing both projection-based metrics and set-based metrics is essential for improving the quality of the reconstruction. However, being effective in projection-based metrics does not necessarily result in being valuable in the context of the set-based metrics.\\
In our recent work \cite{linda}, we investigated the trainability of the projection-dependent detectability index (PDI) for a specific class of objects and its usability for CT trajectory optimization. Because this does not ensure data variability if computed per projection, we additionally introduced a haversine distance constraint in the optimization. The aim is to optimize the quality of the reconstruction in a predefined region of interest. The optimization problem is formulated to maximize the detectability index predicted by a neural network, while ensuring that the haversine distance constraint is met. However, this approach may excluded valuable projections that enhance resolution in the region of interest. \\
In this work, we present an approach to address the issue of manually incorporating data completeness into the data analysis process. Our solution involves integrating this constraint into the neural network architecture. The network output directly indicates an optimized set of projections, which can be used to reconstruct the volume with high accuracy in the region of interest. To achieve this, we propose an adaptation of the ResNet-18 architecture. This outputs a hidden representation for each projection, which are processed through a differentiable ranking function to rank the projections. A straight-through estimator is used to make the final selection of projections. To connect this output to a set of projections, the integer program introduced in \cite{linda} is used to generate a label during training. Our approach is unique in that it directly balances individual projection value and overall set value, ensuring that the final selection of projections maximizes reconstruction quality in the region of interest. We demonstrate the utility of our method in a non-destructive testing scenario by maximizing the reconstruction quality within a specified region of interest using a limited number of projections. \\
To summarize, this paper makes the following contributions:
\begin{itemize}[nosep]
	\item We introduce a network architecture that integrates both projection-based and set-based evaluation metrics for selecting a pre-defined number of projections.
	\item Our approach demonstrates the capability of learning a heuristic that is not limited to individual projections, thereby establishing the basis for using reconstruction-based loss functions to guide the selection of projections in future studies.
\end{itemize}

\section{Methods}
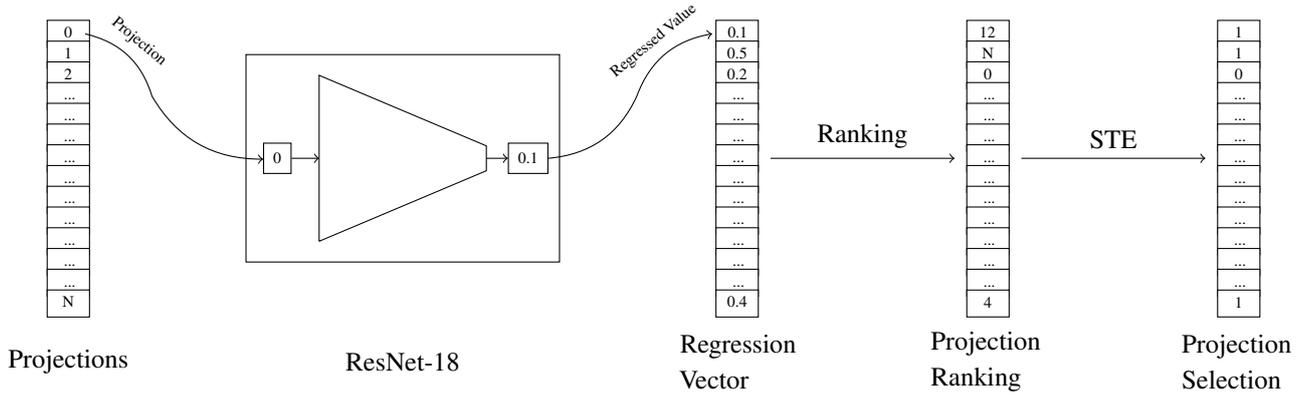
\begin{figure*}
	\centering
\begin{tikzpicture}[scale=1.1]
	\definecolor{fillcolor}{RGB}{255,255,255}
	\newcounter{x}
	\setcounter{x}{0}
	\foreach \y in {3.25,3.0,2.75}{
		\node[rectangle, draw, fill = fillcolor, text width = 0.3cm,text height = 0.1cm, align = center] at(-1., \y){\tiny \arabic{x} };
		\addtocounter{x}{1}
	}
	\foreach \y in {2.5,2.25,2.0,...,0.25}{
		\node[rectangle, draw, fill = fillcolor, text width = 0.3cm,text height = 0.1cm, align = center] at(-1., \y){\tiny ... };
	}
	\node[rectangle, draw, fill = fillcolor, text width = 0.3cm,text height = 0.1cm, align = center] at(-1., 0.0){\tiny N };

	\node[] at(-1.,-0.7) {\small Projections};
	
	\node[rectangle, draw, fill = fillcolor, ] (a) at(1.5, 1.75){\tiny 0};
	\draw[fill = fillcolor] (a) edge[->] (2,1.75);
	\fill[fillcolor, draw = black] (2, 0.75) -- (2,2.75) -- (4,1.9) -- (4,1.6) -- (2,0.75);
	\node[rectangle, draw, fill = fillcolor, ] (b) at(4.5, 1.75){\tiny 0.1};
	\draw[fill = fillcolor] (4,1.75) edge[->] (b);
	\draw[] (1.125, 0.5) rectangle (4.875, 3);
	\draw[->] (b) to[bend right] (5.8,2.5) to[bend left, ->]node[above, rotate = 40]{\tiny Regressed Value} (6.7, 3.25);
	\draw[->] (-0.8, 3.25) to[bend left]node[above, rotate = -40]{\tiny Projection} (0.0,2.5) to[bend right, ->] (a);
	\node[] at(3,-0.7) {\small ResNet-18};
	
	\node[rectangle, draw, fill = fillcolor, text width = 0.3cm,text height = 0.1cm, align = center] at(7., 3.25){\tiny 0.1 };
	\node[rectangle, draw, fill = fillcolor, text width = 0.3cm,text height = 0.1cm, align = center] at(7., 3.0){\tiny 0.5 };
	\node[rectangle, draw, fill = fillcolor, text width = 0.3cm,text height = 0.1cm, align = center] at(7., 2.75){\tiny 0.2 };
	
	\foreach \y in {2.5,2.25,2.0,...,0.25}{
		\node[rectangle, draw, fill = fillcolor, text width = 0.3cm,text height = 0.1cm, align = center] at(7., \y){\tiny ... };
	}
	\node[rectangle, draw, fill = fillcolor, text width = 0.3cm,text height = 0.1cm, align = center] at(7., 0.0){\tiny 0.4 };
	\node[text width = 1.5cm] at(7.,-0.7) {\small Regression\newline Vector};
	
	\draw[fill = fillcolor] (7.4,1.75) edge[->] node[above]{\small Ranking} (9.6,1.75);
	\node[rectangle, draw, fill = fillcolor, text width = 0.3cm,text height = 0.1cm, align = center] at(10., 3.25){\tiny 12 };
	\node[rectangle, draw, fill = fillcolor, text width = 0.3cm,text height = 0.1cm, align = center] at(10., 3.0){\tiny N };
	\node[rectangle, draw, fill = fillcolor, text width = 0.3cm,text height = 0.1cm, align = center] at(10., 2.75){\tiny 0 };
	
	\foreach \y in {2.5,2.25,2.0,...,0.25}{
		\node[rectangle, draw, fill = fillcolor, text width = 0.3cm,text height = 0.1cm, align = center] at(10., \y){\tiny ... };
	}
	\node[rectangle, draw, fill = fillcolor, text width = 0.3cm,text height = 0.1cm, align = center] at(10., 0.0){\tiny 4 };
	
	\node[text width = 1.5cm] at(10,-0.7) {\small Projection\newline Ranking};
	
	\draw[fill = fillcolor] (10.4,1.75) edge[->] node[above]{\small STE} (12.6,1.75);
	\node[rectangle, draw, fill = fillcolor, text width = 0.3cm,text height = 0.1cm, align = center] at(13., 3.25){\tiny 1 };
	\node[rectangle, draw, fill = fillcolor, text width = 0.3cm,text height = 0.1cm, align = center] at(13., 3.0){\tiny 1 };
	\node[rectangle, draw, fill = fillcolor, text width = 0.3cm,text height = 0.1cm, align = center] at(13., 2.75){\tiny 0 };
	
	\foreach \y in {2.5,2.25,2.0,...,0.25}{
		\node[rectangle, draw, fill = fillcolor, text width = 0.3cm,text height = 0.1cm, align = center] at(13., \y){\tiny ... };
	}
	\node[rectangle, draw, fill = fillcolor, text width = 0.3cm,text height = 0.1cm, align = center] at(13., 0.0){\tiny 1 };
	\node[text width = 1.5cm] at(13,-0.7) {\small Projection\newline Selection};
\end{tikzpicture}
\caption{Overview of the proposed approach. First, each projection is regressed to a single value. This value is used as a basis for the ranking. To generate a selection out of the ranking, we utilize a straight trough estimator.}
\label{fig:screenshot001}
\end{figure*}

We propose a combination of projection-based and set-based evaluation metrics using a three-step neural network approach. The first step involves reducing each projection to a single value using a modified ResNet-18 architecture. The regressed values are then collected in a single vector, which represents the valuability of each projection. The second step involves applying a differentiable ranking function (described in Section \ref{ranking}) to this vector, resulting in a ranking of the projections. Lastly, the threshold function of the Straight-Through Estimator (detailed in Section \ref{ste}) is applied to the ranking to convert it into a binary vector that represents the selection of projections. An overview can be seen in Figure \ref{fig:screenshot001}.
\subsection{Projection-Dependent Detectability Index} \label{decidx}
The projection-dependent detectability index is a measure of the quality of a single projection and its ability to contribute to the observability of a signal in a reconstructed image. It is used to evaluate the performance of different CT projection angles. The PDI is calculated using the non-prewhitening matched filter observer model (NPWM) described in Stayman et al. \cite{Thies,Nico,Stayman}. The NPWM model defines the modulation transfer function (MTF) and noise power spectrum (NPS) as functions of the position of the target voxels in the volume, denoted by (x,y,z). The analytical equations for both the MTF and NPS in the context of iterative penalized-likelihood reconstruction were developed by Gang et al. \cite{Gang2014}. The PDI is given by the equation
{\small \begin{equation}
		d^2{(x,y,z)} = \frac{[ \int\!\!\int\!\!\int \lvert MTF(x,y,z) \rvert^2 \lvert W_{t}\rvert^2 df_x df_y df_z ]^2} {\int\!\!\int\!\!\int \lvert NPS(x,y,z) MTF(x,y,z) \rvert^2 \lvert W_{t}\rvert^2 df_x df_y df_z}  \label{dec}
\end{equation}} \hspace*{-5.5pt}
where $ W_{t} $ is the Fourier transform of the region of interest to be imaged with the highest quality. A high detectability index indicates that the required signal is effectively detected in the respective projection according to the task function $ W_{t} $.

\subsection{Differentiable Ranking}\label{ranking}
The ranking operator is a discontinuous function, lacking differentiability, making it unsuitable for use as a component in the training of neural networks, as the gradients necessary for optimization are not defined. To address this issue, differentiable soft ranking, a technique proposed by Blondel et al. \cite{Blondel} allows neural networks to learn a ranking function that can be used to order elements in a set, such as images in a dataset, based on some criterion. The algorithm, a variation of sorting networks, is more efficient in terms of computation time and memory requirements compared to other differentiable sorting algorithms. \\
The basic idea of the technique is to reformulate the descending ranking operation as a linear program over the permutahedron. A permutahedron $P(w)$, where $w \in \mathbb{R}^{n}$, is a polytope that represents the symmetries of a permutation $w$, with vertices corresponding to all possible permutations of a set of elements and edges corresponding to transpositions of adjacent elements. The linear program can be written as
\begin{equation}
	r(\theta) = \underset{\mu \in P({\varphi})}{\arg\max} \ \langle \mu, - \theta \rangle
\end{equation}
where $ \theta \in \mathbb{R}^{n} $ is the vector of scores produced by the neural network and $ {\varphi} = (n,n-1, \dots, 1) $.  An optimal solution is mostly achieved at a vertex of the permutahedron according to the fundamental theorem of linear programming \cite{Danzig}.\\
To ensure differentiability, the authors introduce strongly convex regularization $\Psi$ to the linear program $P_{\Psi}(- \theta, \varphi)$. The regularization strength is controlled through a parameter $\epsilon > 0$, which is multiplied by $\Psi$. The resulting $\Psi$-regularized soft ranking can be formulated as
\begin{equation}
	r_{\epsilon\Psi}(\theta) = P_{\Psi}(\sfrac{- \theta}{\epsilon}, \varphi) 
\end{equation}
In this context, we use a quadratic regularization term $ Q(\mu) = \frac{1}{2} \lVert \mu \rVert^{2} $. Using $ \Psi = Q $, the linear program for the soft ranking over the permutahedron results in
\begin{equation}
	r_{\epsilon Q}(\theta) = P_{Q}(\sfrac{- \theta}{\epsilon}, \varphi) = \underset{\mu \in P({\varphi})}{\arg\max} \ \langle \mu, \sfrac{- \theta}{\epsilon} \rangle - Q(\mu)
\end{equation}
Ascending-order soft ranking can be obtained by negating the input.

\subsection{Straight-Through Estimator}\label{ste}
After obtaining a ranking for the projections, the next step is to convert it into a selection of the projections. To accomplish this task, it is necessary to find a function that assigns a value of 1 to the chosen projections based on the ranking and a value of 0 to the remaining projections. The threshold function, as shown in Equation \eqref{tresh} , is one way to implement this.
\begin{equation}
	thresh(x) = \begin{cases}
		1 & x \le k\\
		0 & x > k
	\end{cases} \label{tresh}
\end{equation}
Here, it is assumed that the ranking is in descending order and the goal is to select k projections. Thresholding, however, often causes issues during backpropagation because its derivative is zero. To address this issue, the Straight-Through Estimator (STE) is used in the backward pass as introduced in \cite{ste}. The STE allows the network to learn from the chosen projections during training by backpropagating through the hard threshold function as if it were the identity function. This allows for the gradients to be computed and updated correctly.

\section{Experiments}
\subsection{Data}
In this work, we conducted simulation experiments within the domain of non-destructive testing. The aim was to identify defects in a pre-specified region of interest. To achieve this, three test specimens of dimensions $\SI{10}{cm} \times \SI{8}{cm} \times \SI{8}{cm}$, made of aluminium with a density of $\SI{2.7}{g/cm^3}$, were selected. Six different representations of each test specimen were generated, each with the same size and density, but with variations in the placement of the spherical-shaped defect, with a radius of $\SI{1}{mm}$. As a result, $18$ distinct test specimens were obtained.\\ 
 We defined a geometry for the CT imaging setup by fixing the detector-isocentre distance and the source-detector distance to $\SI{3}{m}$ and $\SI{4}{m}$, respectively. The detector and source were positioned to face each other, which results in restricting the scan positions to a sphere. We parametrized the problem using azimuth angle $\varphi$ and elevation angle $\theta$ using spherical coordinates. Each CT trajectory consisted of a set of pairs $(\varphi_t,\theta_t)$ for $t \in {0, \dots, N}$, where $N$ is the total number of projection images. Initially, we sampled $N=1000$ scan positions on a sphere using a Fibonacci-based sampling for uniform surface coverage \cite{fib}. \\
 The projection data was simulated using the Fraunhofer EZRT simulation software XSimulation with a \SI{225}{keV} polychromatic spectrum. The test specimens were placed at the centre of the world coordinate system. The detector was chosen to have a size of $375 \times 375$ pixels and a pixel pitch of $\SI{400}{\mu m} \times \SI{400}{\mu m}$. The PDI was calculated analytically using Equation \eqref{dec}. This allowed us to determine a set of projections through the integer optimization problem including the haversine distance constraint introduced in \cite{linda}, which resulted in a binary vector $y \in \{0,1\}^N$ where $1$ indicates a chosen projection. In this work, this used as a label for training. In this study, we set the number of possible projections in the optimized CT trajectory to $k=100$.

\subsection{Neural Network Architecture}
The network was trained on 1000 projections of 15 distinct test specimens, such that 3 test specimens, one for each type of test specimen, remained to test the network performance.
To predict the optimal set of projections for a specific task, we employed a Convolutional Neural Network (CNN) approach. The ResNet-18 architecture was adapted for regression by adding a single fully connected layer to perform the regression towards a scalar value. The network was trained using a Binary Cross Entropy loss function with the Adam optimizer and a learning rate of $ 3{\rm e}^{-5} $. The pre-trained ResNet-18 was initialized with weights from the ImageNet dataset and the fully connected regression layer was initialized randomly. The final selection of projections was determined using the straight-through estimator as described in Section \ref{ste}, yielding a binary representation of the ranking where a value of 1 indicates the projection belongs to the $k$ highest-ranked projections. The training process utilized a batch size of 1 corresponding to a complete CT scan of a test specimen and was performed for a maximum of 600 epochs.

\section{Results}
\begin{figure*}[tbh]
	\centering
	\includegraphics[width=0.9\linewidth]{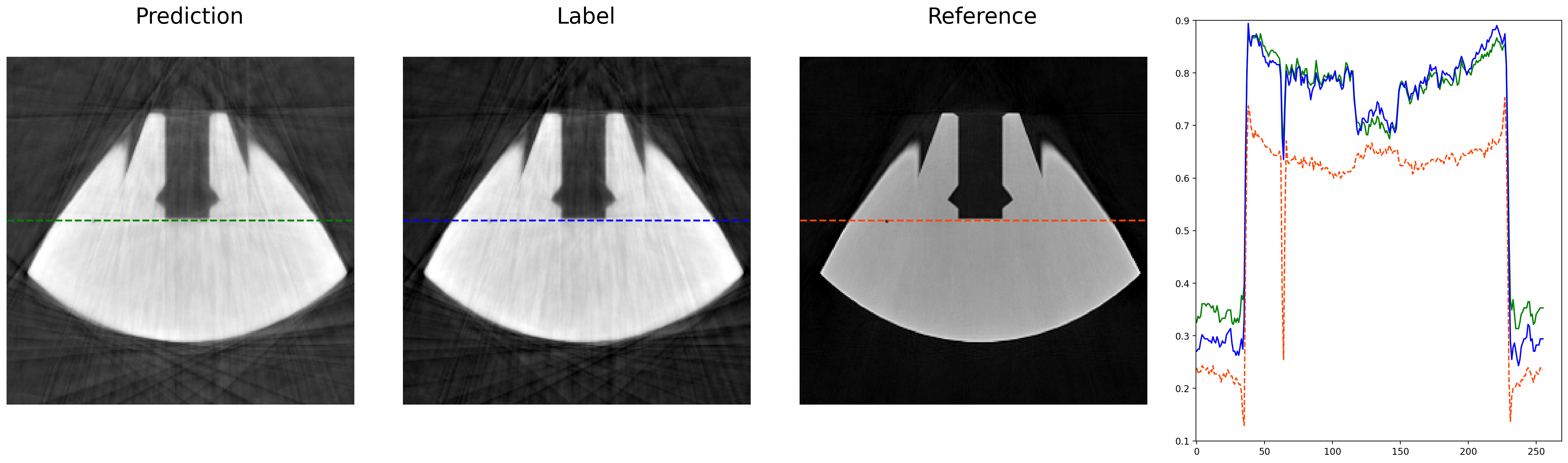}
	\caption{Comparison of the reconstruction quality for the heart gear between the proposed method and the method used to generate the label. Additionally, we show the reference reconstruction generated from the full projection set. One can see, that the proposed method and the labelling method do not show any significant differences, especially in the region around the error. The images are normalized.}
	\label{fig:lineplot}
\end{figure*}
To assess the performance of the proposed method, a comparison was conducted between the reconstructed volume obtained from the predicted optimal set of projections and the optimal set of projections according to the integer program in \cite{linda}. The reconstruction was performed using the Algebraic Reconstruction Technique (ART) with 3 iterations. The evaluation of the reconstructed images was based on two standard image quality metrics,the Structural Similarity Index (SSIM) and the Root Mean Squared Error (RMSE). The results of this comparison are presented in Table \ref{tabularRSME}. Our analysis showed a positive correlation between a decrease in RMSE and an improvement in SSIM. Moreover, the location of the defect was crucial for both the reference method and the proposed method to perform effectively. Our proposed method was able to achieve results that were comparable to the projection set indicated by the label, especially in terms of reconstruction quality.
\begin{table}[htb]
	\centering
	\begin{tabular}[t]{lcccc}
		\toprule
		&\multicolumn{2}{c}{RMSE} &\multicolumn{2}{c}{SSIM}   \\		
		& label & prediction&label& prediction  \\
		\midrule
		Heartgear& 0.16  & 0.13 & 0.73 & 0.79\\ 
		Round Heartgear& 0.21 & 0.21 & 0.45 & 0.48\\ 
		Dent Gear& 0.33 & 0.32 & 0.36 & 0.35 \\ 
		\bottomrule
	\end{tabular}
\caption{Comparison of reconstruction quality by measuring the SSIM and RMSE between the reconstruction of the full projection set and the projections chosen by our approach or the approach in \cite{linda} at the specific region of interest $W_{t}$. We observe that the proposed method performs comparable to the reference method.} \label{tabularRSME}
\end{table}
This is shown Figure \ref{fig:lineplot}, where the grey values of a specific slice of our heartgear test specimen were examined.It was found that the areas of the defect between the prediction and the label are very similar. In addition, it can be seen that the structure of the defect in particular could be clearly depicted, but the same resolution could not be achieved as with the reference reconstruction. Even though there were more reconstruction artefacts in our proposed method, this did not affect the reconstruction quality in the region of interest, as shown by the results in Table \ref{tabularRSME}.

\section{Discussion and Conclusion}
This paper presents a new approach for selecting valuable projections in computed tomography scans, with the aim of improving reconstruction quality in a region of interest. The proposed solution uses a neural network architecture that integrates the projection-based detectability index and data completeness into a single model. The network, based on a modified ResNet-18 architecture, evaluates the value of projections, passes them through a differentiable ranking function, and the final selection is made using a straight-through estimator.\\
The results suggest that our proposed method is capable of selecting projections for a predefined task and a known object structure.
It was found that the performance of the proposed solution is comparable to the baseline method, which is used for labelling. The fact that the performance is comparable to the baseline method demonstrates that the selection of projections can be learned by the neural network. This finding suggests that other set-based metrics, such as a reconstruction-based loss, can also be learned with our presented approach.\\
One drawback of the proposed solution is that it is object-specific, meaning that it is trained on a specific object structure and does not generalize to structures not seen during training. Despite this, the approach is still suitable for non-destructive testing, as it is often necessary to examine similar or identical objects.\\
For future work, we want to add the embedding of positions, which may be useful for improvement. This can help to further guide the neural network in the direction of data completeness.
\section*{Acknowledgements}
This research was financed by the \glqq SmartCT – Methoden der Künstlichen Intelligenz für ein autonomes Roboter-CT System\grqq \ project (project nr. DIK-2004-0009).

\printbibliography

\end{document}